\documentclass[journal]{IEEEtran}
\usepackage{graphicx}
\usepackage[normalem]{ulem}
\usepackage{bm} 
\usepackage{multirow}
\usepackage{amsfonts}
\usepackage{colortbl}
\newcounter{example}

\newcommand{\ignore}[1]{}
\pdfoutput = 1
\begin{document}

\title{Subspace Learning Machine (SLM): \\
Methodology and Performance}
\author{Hongyu~Fu, Yijing~Yang,~\IEEEmembership{Student~Member,~IEEE}
Vinod~K.~Mishra, and C.-C.~Jay~Kuo,~\IEEEmembership{Fellow,~IEEE}%
\thanks{Hongyu~Fu, Yijing~Yang and C.-C.~Jay~Kuo are with Ming Hsieh
Department of Electrical and Computer Engineering, University of
Southern California, Los Angeles, CA 90089, USA, e-mails: hongyufu@usc.edu
(Hongyu Fu), yijingya@usc.edu (Yijing Yang) and cckuo@ee.usc.edu (C.-C. Jay Kuo).}
\thanks{Vinod~K.~Mishra is with Army Research Laboratory, Adelphi, Maryland, USA, e-mail: 
vinod.k.mishra.civ@army.mil (Vinod~K.~Mishra)}
\thanks{This material is based on research sponsored by US Army Research
Laboratory (ARL) under contract number W911NF2020157.}%
}%

\maketitle

\begin{abstract}

Inspired by the feedforward multilayer perceptron (FF-MLP), decision
tree (DT) and extreme learning machine (ELM), a new classification
model, called the subspace learning machine (SLM), is proposed in this
work. SLM first identifies a discriminant subspace, $S^0$, by examining
the discriminant power of each input feature. Then, it uses
probabilistic projections of features in $S^0$ to yield 1D subspaces and
finds the optimal partition for each of them. This is equivalent to
partitioning $S^0$ with hyperplanes. A criterion is developed to choose
the best $q$ partitions that yield $2q$ partitioned subspaces among
them.  We assign $S^0$ to the root node of a decision tree and the
intersections of $2q$ subspaces to its child nodes of depth one. The
partitioning process is recursively applied at each child node to build
an SLM tree. When the samples at a child node are sufficiently pure, the
partitioning process stops and each leaf node makes a prediction. The
idea can be generalized to regression, leading to the subspace learning
regressor (SLR). Furthermore, ensembles of SLM/SLR trees can yield a
stronger predictor. Extensive experiments are conducted for performance
benchmarking among SLM/SLR trees, ensembles and classical
classifiers/regressors. 

\end{abstract}

Machine Learning, Classification, Subspace Learning

\section{Introduction}\label{sec:introduction}

Feature-based classification models have been well studied for many
decades.  Feature extraction and classification are treated as two
separate modules in the classical setting.  Attention has been shifted
to deep learning (DL) in recent years. Feature learning and
classification are handled jointly in DL models. Although the best
performance of classification tasks is broadly achieved by DL through
back propagation (BP), DL models suffer from lack of interpretability,
high computational cost and high model complexity.  Under the classical
learning paradigm, we propose new high-performance classification and
regression models with features as the input in this work. 

Examples of classical classifiers/regressors include support vector
machine (SVM) \cite{svm}, decision tree (DT) \cite{CART}, multilayer
perceptron (MLP) \cite{rosenblatt1958perceptron}, feedforward multilayer
perceptron (FF-MLP) \cite{ruiyuan} and extreme learning machine (ELM)
\cite{ELM}.  SVM, DT and FF-MLP share one common idea, i.e., feature
space partitioning. Yet, they achieve this objective by different means.
SVM partitions the space by leveraging kernel functions and support
vectors.  DT partitions one space into two subspaces by selecting the
most discriminant feature one at a time recursively. DT tends to overfit
the training data when the tree depth is high. To avoid it, one can
build multiple DTs, each of which is a weak classifier, and their
ensemble yields a strong one, e.g., the random forest (RF) classifier
\cite{RF}.  Built upon linear discriminant analysis, FF-MLP uses the
Gaussian mixture model (GMM) to capture feature distributions of
multiple classes and adopts neuron pairs with weights of opposite
signed vectors to represent partitioning hyperplanes. 

The complexity of SVM and FF-MLP depends on the sample distribution in
the feature space. It is a nontrivial task to determine suitable
partitions when the feature dimension is high and/or the sample
distribution is complicated. These challenges could limit the
effectiveness and efficiency of SVM and FF-MLP. Selecting a partition in
DTs is easier since it is conducted on a single feature. Yet, the
simplicity is paid by a price. That is, the discriminant power of an
individual feature is weak, and a DT results in a weak classifier.
As proposed in ELM \cite{ELM}, another idea of subspace partitioning is
to randomly project a high-dimensional space to a 1D space and find the
optimal split point in the associated 1D space.  Although ELM works
theoretically, it is not efficient in practice if the feature dimension
is high. It takes a large number of trials and errors in finding good
projections. A better strategy is needed.

By analyzing pros and cons of SVM, FF-MLP, DT and ELM, we attempt to
find a balance between simplicity and effectiveness and propose a new
classification-oriented machine learning model in this work. Since it
partitions an input feature space into multiple discriminant subspaces
in a hierarchical manner, it is named the subspace learning machine
(SLM). Its basic idea is sketched below. Let $X$ be the input feature
space.  First, SLM identifies subspace $S^0$ from $X$.  If the dimension
of $X$ is low, we set $S^0=X$. If the dimension of $X$ is high, we
remove less discriminant features from $X$ so that the dimension of
$S^0$ is lower than that of $X$. 

Next, SLM uses probabilistic projections of features in $S^0$ to yield
$p$ 1D subspaces and find the optimal partition for each of them. This
is equivalent to partitioning $S^0$ with $2p$ hyperplanes. A criterion
is developed to choose the best $q$ partitions that yield $2q$
partitioned subspaces among them.  We assign $S^0$ to the root node of a
decision tree and the intersections of $2q$ subspaces to its child nodes
of depth one.  The partitioning process is recursively applied at each
child node to build an SLM tree until stopping criteria are met, then
each leaf node makes a prediction. Generally, an SLM tree is wider and
shallower than a DT.  The prediction capability of an SLM tree is
stronger than that of a single DT since it allows multiple decisions at
a decision node.  Its performance can be further improved by ensembles
of multiple SLM trees obtained by bagging and
boosting.  The idea can be generalized to regression, leading to the
subspace learning regressor (SLR).  Extensive experiments are conducted
for performance benchmarking among individual SLM/SLR trees, multi-tree
ensembles and several classical classifiers and regressors. They show
that SLM and SLR offer light-weight high-performance classifiers and
regressors, respectively. 

The rest of this paper is organized as follows. The SLM model is
introduced in Sec.  \ref{sec:method}.  The ensemble design is proposed
in Sec. \ref{sec:ensemble}. Performance evaluation and benchmarking of
SLM and popular classifiers are given in Sec.  \ref{sec:experiments}.
The generalization to SLR is discussed in Sec.  \ref{sec:SLR}. The
relationship between SLM/SLR and other machine learning methods such as
classification and regression tree (CART), MLP, ELM, RF and Gradient
Boosting Decision Tree (GBDT) \cite{GBDT} is described in Sec.
\ref{sec:review}.  Finally, concluding remarks are given in Sec.
\ref{sec:conclusion}. 

\section{Subspace Learning Machine (SLM)}\label{sec:method}

\subsection{Motivation}

Consider an input feature space, $X$, containing $L$ samples, where
each sample has a $D$-dimensional feature vectors. A sample is denoted by
\begin{equation}\label{eq:sample}
\bm{x}_{l} = (x_{l,1} \cdots x_{l,d} \cdots, x_{l,D})^T \in \mathbb{R}^{D},
\end{equation}
where $l = 1, \cdots, L$. We use $F_d$ to represent the $d$th feature 
set of $\bm{x}_{l}$, i.e.,
\begin{equation}\label{eq:Fd}
F_d = \{ x_{l,d} \mid  1 \le l \le L \}.
\end{equation}
For a multi-class classification problem with $K$ classes, each training
feature vector has an associated class label in form of one-hot vector
\begin{equation}\label{eq:label_1}
\bm{y}_l = (y_{l,1}, \cdots, y_{l,k}, \cdots, y_{l,K})^T \in 
\mathbb{R}^{K},
\end{equation}
where
\begin{equation}\label{eq:label_2}
y_{l,k}=1 \mbox{   and   } y_{l,k'}=0, k' \neq k,
\end{equation}
if the $l$th sample belongs to class $k$, where $1 \le k, k' \le K$.
Our goal is to partition the feature space, $\mathbb{R}^{D}$, into
multiple subspaces hierarchically so that samples at leaf nodes are as
pure as possible. It means that the majority of samples at a node is
from the same class. Then, we can assign all samples in the leaf node to
the majority class.  This process is adopted by a DT classifier.  The
root node is the whole sample space, and an intermediate or leaf node
corresponds to a partitioned subspace.  We use $S^0$ to represent the
sample space at the root node and $S^m$, $m=1, \cdots, M$, to denote
subspaces of child nodes of depth $m$ in the tree. 

\begin{figure}[t]
\begin{center}
\includegraphics[width=0.7\linewidth]{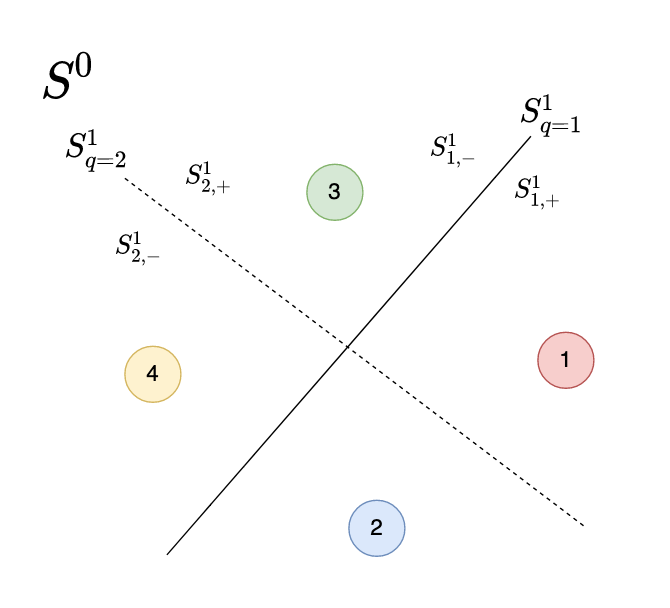} \\ (a) \\
\includegraphics[width=0.7\linewidth]{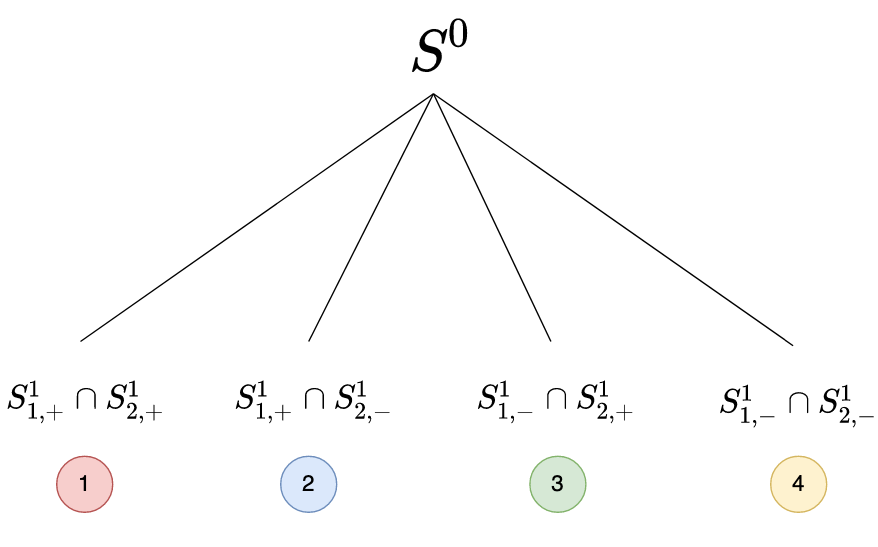} \\ (b)
\end{center}
\caption{(a) An illustration of SLM, where space $S^0$ is partitioned 
into 4 subspaces with two splits, and (b) the corresponding SLM tree 
with a root note and four child nodes.}\label{fig:1}
\end{figure}

The efficiency of traditional DT methods could be potentially improved 
by two ideas. They are elaborated below.
\begin{enumerate}
\item Partitioning in flexibly chosen 1D subspace: \\
We may consider a general projected 1D subspace defined by
\begin{equation}\label{eq:gp}
F_{\bm{a}} = \{f(\bm{a}) \mid f(\bm{a})=\bm{a}^T \bm{x} \},
\end{equation}
where 
\begin{equation}\label{eq:a}
\bm{a}=(a_1, \cdots, a_d, \cdots, a_D)^T, \hspace{5mm} || \bm{a} ||=1,
\end{equation}
The DT is a special case of Eq. (\ref{eq:gp}), where $\bm{a}$ is set to
the $d$th basis vector, $\bm{e}_d$, $1 \le d \le D$. On one hand, this
choice simplifies computational complexity, which is particularly
attractive if $D >> 1$. On the other hand, if there is no discriminant
feature $F_d$, the decision tree will not be effective. It is desired to
find a discriminant direction, $\bm{a}$, so that the subspace,
$F_{\bm{a}}$, has a more discriminant power at a low computational cost. 
\item N-ary split at one node: \\
One parent node in DT is split into two child nodes. We may allow an
n-ary split at the parent. One example is shown in Fig.  \ref{fig:1}(a)
and (b), where space $S^0$ is split into four disjoint subspaces.
Generally, the n-ary split gives wider and shallower decision trees so
that overfitting can be avoided more easily. 
\end{enumerate}
The SLM method is motivated by these two ideas. Although the high-level
ideas are straightforward, their effective implementations are
nontrivial. They will be detailed in the next subsection. 

\subsection{Methodology}\label{subsec:RandomProjection}

Subspace partitioning in a high-dimensional space plays a fundamental
role in the design of powerful machine learning classifiers. Generally,
we can categorize the partitioning strategies into two types: 1) search
for an optimal split point in a projected 1D space (e.g., DT) and 2)
search for an optimal splitting hyperplane in a high-dimensional space
(e.g., SVM and FF-MLP). Mathematically, both of them can be expressed in
form of
\begin{equation}\label{eq:hyper-plane}
\bm{a}^T \bm{x} - b =0,
\end{equation}
where $-b$ is called the bias and 
\begin{equation}\label{eq:orientation}
\bm{a}=(a_1, \cdots, a_d, \cdots, a_D)^T, \hspace{5mm} || \bm{a} ||=1,
\end{equation}
is a unit normal vector that points to the surface normal direction.  It
is called the projection vector. Then, the full space, $S$, is split into 
two half subspaces:
\begin{equation}\label{eq:split}
S_+: \, \bm{a}^T \bm{x} \ge b, \mbox{   and   } S_-: \, \bm{a}^T \bm{x} < b.
\end{equation}

The challenge lies in finding good projection vector $\bm{a}$ so that
samples of different classes are better separated. It is related to the
distribution of samples of different classes.  For the first type of
classifiers, they pre-select a set of candidate projection vectors, try
them out one by one, and select the best one based on a certain
criterion. For the second type of classifiers, they use some criteria
to choose good projection vectors. For example, SVM first identifies
support vectors and then finds the hyperplane that yields the largest
separation (or margin) between two classes.  The complexity of the first
type is significantly lower than that of the second type. 

In SLM, we attempt to find a mid-ground of the two. That is, we
generate a new 1D space as given by
\begin{equation}\label{eq:gp2}
F_{\bm{a}} = \{f(\bm{a}) \mid f(\bm{a})=\bm{a}^T \bm{x} \},
\end{equation}
where $\bm{a}$ is a vector on the unit hypersphere in $\mathbb{R}^{D}$
as defined in Eq. (\ref{eq:orientation}). By following the first type of
classifier, we would like to identify a set of candidate projection
vectors. Yet, their selection is done in a probabilistic manner.
Generally, it is not effective to draw $\bm{a}$ on the unit hypersphere
uniformly. The criterion of effective projection vectors and their
probalisitic selection will be presented in Secs.
\ref{sec:good_projection}-\ref{subsubsec:Uncorrelatedness}.  Without
loss of generality, we primarily focus on the projection vector
selection at the root node in the following discussion. The same idea
can be easily generalized to child nodes. 


\subsubsection{Selection Criterion}\label{sec:good_projection}

We use the discriminant feature test (DFT) \cite{yang2022supervised} to
evaluate the discriminant quality of the projected 1D subspace as given
in Eq. (\ref{eq:gp2}). It is summarized below.

For a given projection vector, $\bm{a}$, we find the minimum and the
maximum of projected values $f(\bm{a})=\bm{a}^T \bm{x}$, which are
denoted by $f_{\min}$ and $f_{\max}$, respectively. We partition
interval $[f_{\min}, f_{\max}]$ into $B$ bins uniformly and use bin
boundaries as candidate thresholds. One threshold, $t_b$, $b=1,
\cdots, B-1$, partitions interval $[f_{\min}, f_{\max}]$ into 
two subintervals that define two sets:
\begin{eqnarray}
F_{\bm{a},t_b,+} &=&  \{f(\bm{a})  \mid \bm{a}^T \bm{x} \ge t_b \}, \\
F_{\bm{a},t_b,-} &=&  \{f(\bm{a})  \mid \bm{a}^T \bm{x} < t_b  \}.
\end{eqnarray}
The bin number, $B$, is typically set to 16 \cite{yang2022supervised}.

The quality of a split can be evaluated with the weighted sum of loss 
functions defined on the left and right subintervals:
\begin{eqnarray}\label{eq:weighted_entropy_cost}
\mathcal{L}_{\bm{a},t_b}  =  \frac{N_+}{N_+ + N_-} \mathcal{L}_{\bm{a},t_b,+}
+ \frac{N_-}{N_+ + N_-} \mathcal{L}_{\bm{a},t_b,-},
\end{eqnarray}
where $N_+=\mid F_{\bm{a},t_b,+} \mid$ and $N_-=\mid F_{\bm{a},t_b,-}
\mid$ are sample numbers in the left and right subintervals,
respectively.  One convenient choice of the cost function is the entropy
value as defined by
\begin{eqnarray}\label{eq:entropy_cost}
\mathcal{L} = - \sum^C_{c=1} p_c \log (p_c),
\end{eqnarray}
where $p_c$ is the probability of samples in a certain set belonging to
class $c$ and $C$ is the total number of classes. In practice, the
probability is estimated using the histogram.  Finally, the discriminant
power of a certain projection vector is defined as the minimum cost
function across all threshold values:
\begin{eqnarray}\label{eq:min_entropy_cost}
\mathcal{L}_{\bm{a}, \mbox{opt}}= \min_{t_b} \mathcal{L}_{\bm{a},t_b}.
\end{eqnarray}
We search for projection vectors, $\bm{a}$ that provide small cost
values as give by Eq. (\ref{eq:min_entropy_cost}). The smaller, the
better. 

\subsubsection{Probabilistic Selection}\label{sec:projection}

We adopt a probabilistic mechanism to select one or multiple good
projection vectors. To begin with, we express the project vector as
\begin{equation}\label{eq:basis}
\bm{a}=a_1 \bm{e}_1 + \cdots, a_d \bm{e}_d + \cdots + a_D \bm{e}_D,
\end{equation}
where $\bm{e}_d$, $d=1,\cdots,D$, is the basis vector.  We evaluate the
discriminant power of $\bm{e}_d$ by setting $\bm{a}=\bm{e}_d$ and
following the procedure in Sec. \ref{sec:good_projection}. Our
probabilistic selection scheme is built upon one observation.  The
discriminability of $\bm{e}_d$ plays an important role in choosing a
more discriminant $\bm{a}$. Thus, we rank $\bm{e}_d$ according to their
discriminant power measured by the cost function in Eq.
(\ref{eq:min_entropy_cost}). The newly ordered basis is denoted by
$\bm{e'}_d$, $d=1,\cdots,D$, which satisfies the following:
\begin{eqnarray}\label{eq:new_order}
\mathcal{L}_{\bm{e'}_1, \mbox{opt}} \le \mathcal{L}_{\bm{e'}_2, 
\mbox{opt}} \le \cdots \le \mathcal{L}_{\bm{e'}_D, \mbox{opt}}. 
\end{eqnarray}
We can rewrite Eq. (\ref{eq:basis}) as
\begin{equation}\label{eq:new_basis}
\bm{a}=a'_1 \bm{e'}_1 + \cdots, a'_d \bm{e'}_d + \cdots + 
a'_D \bm{e'}_D.
\end{equation}

We use three hyper-parameters to control the probabilistic selection 
procedure. 
\begin{itemize}
\item $P_d$: the probability of selecting coefficient $a'_d$ \\ Usually,
$P_d$ is higher for smaller $d$. In the implementation, we adopt the
exponential density function in form of
\begin{equation}\label{eq:exp_p}
P_d= \beta_0 \exp(-\beta d), 
\end{equation}
where $\beta > 0$ is a parameter and $\beta_0$ is the corresponding
normalization factor. 
\item $A_d$: the dynamic range of coefficient $a'_d$ \\
To save the computation, we limit the values of $a'_d$ and consider 
integer values within the dynamic range; namely,
\begin{equation}\label{eq:dynamic}
a'_d = 0, \pm 1, \, \pm 2, \, \cdots, \, \pm \lfloor A_d \rfloor,
\end{equation}
where $A_d$ is also known as the envelop parameter. Again, we adopt the
exponential density function 
\begin{equation}\label{eq:exp_a}
A_d= \alpha_0 \exp(-\alpha d), 
\end{equation}
where $\alpha > 0$ is a parameter and $\alpha_0$ is the corresponding
normalization factor in the implementation. When the search space in Eq.
(\ref{eq:dynamic}) is relatively small with chosen hyperparameters, we
exhaustively test all the values. Otherwise, we select a partial set of
the orientation vector coefficients probabilistically under the uniform
distribution. 

\item $R$: the number of selected coefficients, $a'_d$, in Eq
(\ref{eq:new_basis}) \\
If $D$ is large, we only select a subset of $R$ coefficients, where $R << D$ 
to save computation. The dynamic ranges of the remaining $(D-R)$ coefficients
are all set to zero.
\end{itemize}
By fixing parameters $\beta$, $\alpha$ and $R$ in one round of $\bm{a}$
generation, the total search space of $\bm{a}$ to be tested by DFT lies between
\begin{equation}
\mbox{U.B.}= \Pi_{d=1}^R (2 A_d + 1), \quad
\mbox{L.B.}= \Pi_{d=D+1-R}^D (2 A_d + 1),
\end{equation}
where U.B and L.B. mean upper and lower bounds, respectively.  To
increase the diversity of $\bm{a}$ furthermore, we may use multiple rounds in the
generation process with different $\beta$, $\alpha$ and $R$ parameters. 

\begin{figure}[t]
\begin{center}
\includegraphics[width=0.9\linewidth]{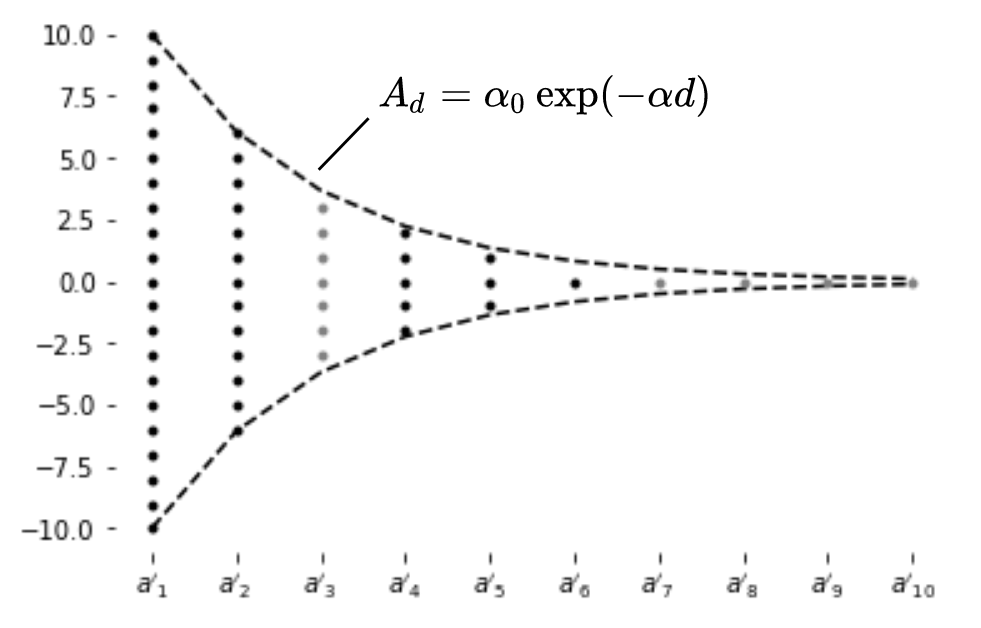}
\end{center}
\caption{Illustration of the probabilistic selection process, where the
envelop function $A_d$ that provides a bound on the magnitude of
coefficients $a'_d$ in the orientation vector. The dimensions with black
dots are selected dimensions, and dots in one vertical line are integers
for selection for each dimension. In this example, the selected
dimensions are $a'_1$, $a'_2$, $a'_4$, $a'_5$ and $a'_6$.  For each
trial, we select one black dot per vertical line to form an orientation
vector. The search can be done exhaustively or randomly with the uniform 
distribution.}\label{fig:envelop}
\end{figure}

We use Fig. \ref{fig:envelop} as an example to illustrate the probabilistic
selection process. Suppose input feature dimension $D=10$ and $R$ is
selected as 5, we may apply $\alpha_0$ as 10 and $\alpha$ as 0.5 for
bounding the dynamic range of the $a'_d$ selections. During the
probabilistic selection, the $R=5$ coefficients are selected with the
candidate integers for the corresponding $a'_d$ marked as black dots,
the unselected $D-R$ coefficients are marked as gray and the actual
coefficients are set to zero. The coefficients of the orientation vector
$\bm{a}$ are uniformly selected among the candidate integers marked as
black dots in Fig. \ref{fig:envelop}. 

\subsubsection{Selection of Multiple Projection Vectors} 
\label{subsubsec:Uncorrelatedness}

Based on the search given in Sec. \ref{sec:projection}, we often 
find multiple effective projection vectors 
\begin{equation}\label{eq:gprojection}
\bm{\tilde{a}}_j ,j=1, \cdots, J,
\end{equation}
which yield discriminant 1D subspaces.  It is desired to leverage them
as much as possible. However, many of them are highly correlated in the
sense that their cosine similarity measure is close to unity, i.e.,
\begin{equation}
\bm{\tilde{a}}_j^T \bm{\tilde{a}}_{j'} \approx 1.
\end{equation}
As a result, the correspondidg two hyperplanes have a small angle
between them. To avoid this, we propose an iterative procedure to select
multiple projection vectors. To begin with, we choose the projection
vector in Eq. (\ref{eq:gprojection}) that has the smallest cost as the
first one. Next, we choose the second one from Eq.
(\ref{eq:gprojection}) that minimizes its absolute cosine similarity
value with the first one. We can repeat the same process by minimizing
the maximum cosine similarities with the first two, etc.  The iterative
mini-max optimization process can be terminated by a pre-defined
threshold value, $\theta_{minimax}$. 

\subsubsection{SLM Tree Construction}\label{subsec:stopping}

\begin{figure}[t]
\begin{center}
\includegraphics[width=0.9\linewidth]{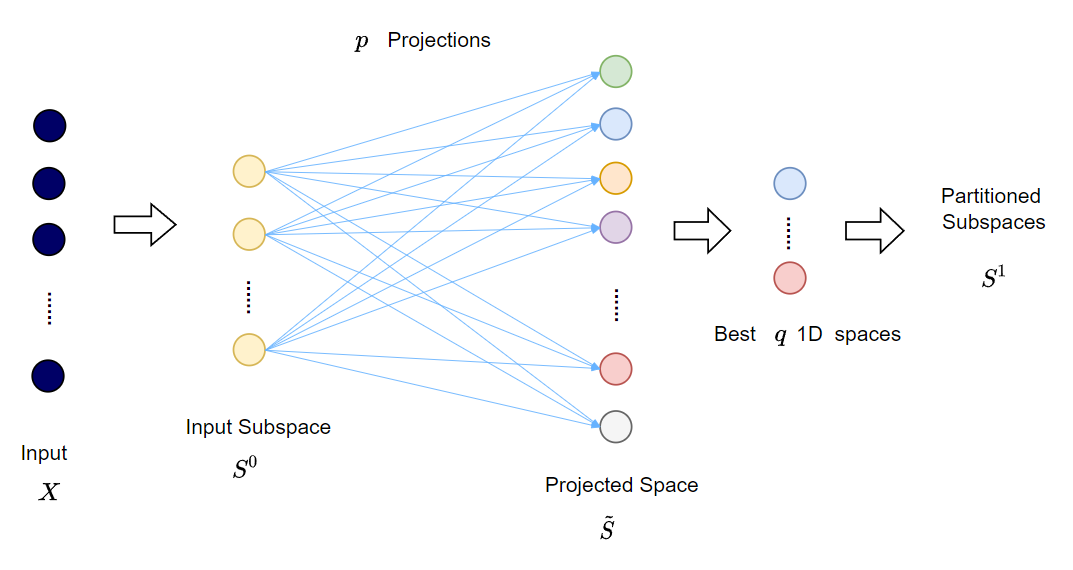}
\end{center}
\caption{An overview of the SLM system.}\label{fig:1_b}
\end{figure}

We illustrate the SLM tree construction process in Fig. \ref{fig:1_b}. 
It contains the following steps.
\begin{enumerate}
\item Check the discriminant power of ${D}$ input dimensions and find
discriminant input subspace ${S}^{0}$ with ${D}_{0}$ dimensions among D. 

\item Generate $p$ projection vectors that project the selected input subspace to $p$ 1D subspaces. The projected space is denoted by $\tilde{S}$. 

\item Select the best $q$ 1D spaces from $p$ candidate subspaces based
on discriminability and correlation and split the node accordingly,
which is denoted by $S^{1}$. 

\end{enumerate}

The node split process is recursively conducted to build nodes of the
SLM tree. The following stopping criteria are adopted to avoid
overfitting at a node.
\begin{enumerate}
\item The depth of the node is greater than user's pre-selected
threshold (i.e.  the hyper-parameter for the maximum depth of an SLM
tree). 
\item The number of samples at the node is less than user's pre-selected
threshold (i.e. the hyper-parameter for the minimum sample number per
node). 
\item The loss at the node is less than user's pre-selection threshold 
(i.e. the hyper-parameter for the minimum loss function per
node). 
\end{enumerate}

\section{SLM Forest and SLM Boost} \label{sec:ensemble}

Ensemble methods are commonly used in the machine learning field to
boost the performance. An ensemble model aims to obtain better
performance than each constituent model alone. With DTs as the
constituent weak learners, the bootstrap aggregating or bagging method,
(e.g., RF) and the boosting method (e.g. GBDT) are the most popular
ensemble methods.  As compared to other classical machine learning
methods, they often achieve better performance when applied to real
world applications. In this section, we show how to apply the ensemble
idea to one single SLM tree (i.e., SLM Baseline).  Inspired by RF, we
present a bagging method for SLM and call it the SLM Forest. Similarly,
inspired by XGBoost \cite{chen2015xgboost, chen2016xgboost}, we propose
a boosting method and name it SLM Boost.  They are detailed in Secs.
\ref{subsec:forest} and \ref{subsec:boosting}, respectively. 

\subsection{SLM Forest}\label{subsec:forest}

For traditional DTs, RF is the most popular bagging ensemble algorithm.
It consists of a set of tree predictors, where each tree is built based
on the values of a random vector sampled independently and with the same
distribution for all trees in the forest \cite{RF}.  With the Strong Law
of Large Numbers, the performance of RF converges as the tree number
increases. As compared to the individual DTs, significant performance
improvement is achieved with the combination of many weak decision
trees.  Motivated by RF, SLM Forest is developed by learning a series of
single SLM tree models to enhance the predictive performance of each
individual SLM tree. As discussed in Sec. \ref{sec:method}, SLM is a
predictive model stronger than DT. Besides, the probabilistic projection
provides diversity between different SLM models.  Following RF, SLM
Forest takes the majority vote of the individual SLM trees as the
ensemble result for classification tasks, and it adopts the mean of each
SLM tree prediction as the ensemble result for regression tasks. 

It is proved in \cite{amit1997shape} that the predictive performance of
RF depends on two key points: 1) the strength of individual trees, and
2) the weak dependence between them. In other words, a high performance
RF model can be obtained through the combination of strong and
uncorrelated individual trees.  The model diversity of RF is achieved
through bagging of the training data and feature randomness.  For the
former, RF takes advantage of the sensitivity of DTs to the data they
are trained on, and allows each individual tree to randomly sample from
the dataset with replacement, resulting in different trees. For the
latter, each tree can select features only from a random subset of
the whole input features space, which forces diversity among trees
and ultimately results in lower correlation across trees for better
performance. 

SLM Forest achieves diversity of SLM trees more effectively through
probabilistic selection as discussed in Sec.  \ref{sec:projection}.  For
partitioning at a node, we allow a probabilistic selection of ${D}_{0}$
dimensions from the $D$ input feature dimensions by taking the
discriminant ability of each feature into account.  In Eq.
(\ref{eq:exp_p}), $\beta$ is a hyper-parameter used to control the
probability distribution among input features. A larger $\beta$ value
has higher preference on more discriminant features. Furthermore, the
envelope vector, $\bm{A}_{d}$ in Eq. (\ref{eq:exp_a}) gives a bound to
each element of the orientation vector. It also attributes to the
diversity of SLM trees since the search space of projection vectors are
determined by hyper-parameter $\alpha$.  Being similar to the
replacement idea in RF, all training samples and all feature dimensions
are kept as the input at each node splitting to increase the strength of
individual SLM trees. 

With individual SLM trees stronger than individual DTs and novel design
in decorrelating partitioning planes, SLM Forest achieves better
performance and faster converge than RF. This claim is supported by
experimental results in Sec. \ref{sec:experiments}. 

\subsection{SLM Boost}\label{subsec:boosting}

With standard DTs as weak learners, GBDT \cite{GBDT} and XGBoost
\cite{chen2015xgboost, chen2016xgboost} can deal with a large amount of
data efficiently and achieve the state-of-the-art performance in many
machine learning problems. They take the ensemble of standard DTs with
boosting, i.e. by defining an objective function and optimizing it with
learning a sequence of DTs.  By following the gradient boosting
process, we propose SLM Boost to ensemble a sequence of SLM trees. 

Mathematically, we aim at learning a sequence of SLM trees, where the
$t$th tree is denoted by $f_{t}({\bf x})$. Suppose that we have $T$
trees at the end. Then, the prediction of the ensemble is the sum of all
trees, i.e. each of the $L$ samples is predicted as
\begin{eqnarray} \label{eqn:xgb_pred}
\hat{y}_{l}  &=& \sum^T_{t=1} f_{t} ({\bf x}_{l}) \quad l = 1,2,\cdots,L.
\end{eqnarray}
The objective function for the first $t$ trees is defined as
\begin{eqnarray} \label{eqn:obj}
\Omega  =  \sum^L_{l=1} \gamma ({y}_{l}, \hat{y}_{l}^{(t)}),
\end{eqnarray}
where $\hat{y}_{l}^{(t)}$ is the prediction of sample $l$ with all $t$
trees and $\gamma({y}_{l}, \hat{y}_{l}^{(t)})$ denotes the training loss
for the model with a sequence of $t$ trees.  The log loss and the mean
squared error are commonly utilized for classification and regression
tasks as the training loss, respectively.  It is intractable to learn
all trees at once. To design SLM Boost, we follow the GBDT process and
use the additive strategy. That is, by fixing what have been learned
with all previous trees, SLM Boost learns a new tree at each time.
Without loss of generality, we initialize the model prediction as 0.
Then, the learning process is
\begin{eqnarray}
\hat{y}_{l}^{(0)} & = & 0 \\
\hat{y}_{l}^{(1)} & = & f_{1}({\bf x}_{l}) = \hat{y}_{l}^{(0)} + f_{1}(\bf{x}_{l})\\
& \cdots & \\
\hat{y}_{l}^{(t)} & =  & \sum^t_{i=1} f_{i}({\bf x}_{l}) = \hat{y}_{l}^{(t-1)} 
+ f_{t}(x_{l})
\end{eqnarray}
Then, the objective function to learn the $t$th tree can be written as
\begin{eqnarray} \label{eqn:obj5}
\Psi(t)  = \sum^L_{l=1} \gamma({y}_{l}, \hat{y}_{l}^{(t-1)} + f_{t}({\bf x}_{l})).
\end{eqnarray}
Furthermore, we follow the XGBoost process and take the Taylor Expansion of the
loss function up to the second order to approximate the loss function in
general cases. Then, the objective function can be approximated as
\begin{eqnarray} \label{eqn:obj2}
\Psi^{(t)}  \approx \sum^L_{l=1} (\gamma({y}_{l}, \hat{y}_{l}^{(t-1)}) + 
g_{l}f_{t}({\bf x}_{l}) + \frac{1}{2}h_{l}f^{2}_{t}({\bf x}_{l})) + C,
\end{eqnarray}
where $g_{l}$ and $h_{l}$ are defined as
\begin{eqnarray}
g_{l} & = & \partial_{\hat{y}_{l}^{(t-1)}} \gamma({y}_{l}, \hat{y}_{l}^{(t-1)})  \label{eqn:obj3} \\
h_{l} & = & \partial^{2}_{\hat{y}_{l}^{(t-1)}} \gamma({y}_{l}, \hat{y}_{l}^{(t-1)})  \label{eqn:obj6}
\end{eqnarray}
After removing all constants, the objective function for the $t$th SLM tree becomes
\begin{eqnarray} \label{eqn:obj4}
\sum^L_{l=1} [g_{l}f_{t}({\bf x}_{l}) + \frac{1}{2}h_{l}f^{2}_{t}({\bf x}_{l})]
\end{eqnarray}
With individual SLM trees stronger than individual DTs, SLM Boost
achieves better performance and faster convergence than XGBoost as
illustrated by experimental results in Sec. \ref{sec:experiments}. 

\section{Performance Evalution of SLM}\label{sec:experiments}

\begin{table*}[!htbp]
  \begin{center}
    \caption{Classification accuracy comparison of 10 benchmarking methods on nine datasets.} 
    \label{table:performance}
     \begin{tabular}{ |l|p{1.25cm}|p{1.25cm}|p{1.25cm}|p{1.25cm}|p{1.25cm}|p{1.25cm}|p{1.25cm}|p{1.5cm}|p{1.25cm}|}
     \hline
      \multirow{2}{*}{\textbf{Datasets}} &  circle-and-ring	& 2-new-moons &	4-new-moons	& Iris & Wine & B.C.W. & Pima & Ionosphere & Banknote \\ \hline
        FF-MLP  & 87.25&	91.25&	95.38&	98.33	&94.44&	94.30&	73.89&	89.36&	98.18\\
        BP-MLP  & 88.00&	91.25&	87.00&	64.67&	79.72&	97.02&	75.54	&84.11&	88.38\\
        LSVM&	48.50 &	85.25&	85.00&	96.67&	\textbf{98.61}&	96.49&	76.43&	86.52&	99.09\\
        SVM/RBF &\textbf{88.25}&	89.75&	88.38&	\textbf{98.33}&	\textbf{98.61}&	\textbf{97.36} 
                &75.15&	\textbf{93.62} & \textbf{100.00}\\
        DT &	85.00 &	87.25&	94.63&	\textbf{98.33}&	95.83&	94.74 &	77.07&	89.36&	98.00\\
        SLM Baseline & \textbf{88.25}	& \textbf{91.50}	&\textbf{95.63}	&\textbf{98.33}& 
        \textbf{98.61} & 97.23 & \textbf{77.71}  & 90.07& 99.09\\ \hline \hline
        RF &	87.00   &	90.50&	\textbf{96.00}&	\textbf{98.33}&	\textbf{100.00}&	95.61&	\textbf{79.00}&	94.33&	98.91\\
        XGBoost&	87.50&	91.25&	\textbf{96.00}&	\textbf{98.33}&	\textbf{100.00}	&98.25&	75.80&	91.49&	99.82\\
        SLM Forest	&\textbf{88.25}&	\textbf{91.50}&	\textbf{96.00} &\textbf{98.33}&	\textbf{100.00}	&97.36&	\textbf{79.00}&	\textbf{95.71}&	\textbf{100.00}\\
        SLM Boost	&\textbf{88.25}	&\textbf{91.50}&	\textbf{96.00}	&\textbf{98.33}&	\textbf{100.00}	&\textbf{98.83}&	77.71&	94.33&	\textbf{100.00}\\ \hline
    \end{tabular}
  \end{center}
\end{table*}

\begin{figure*}[!t]
\centering
\includegraphics[width=0.9\linewidth]{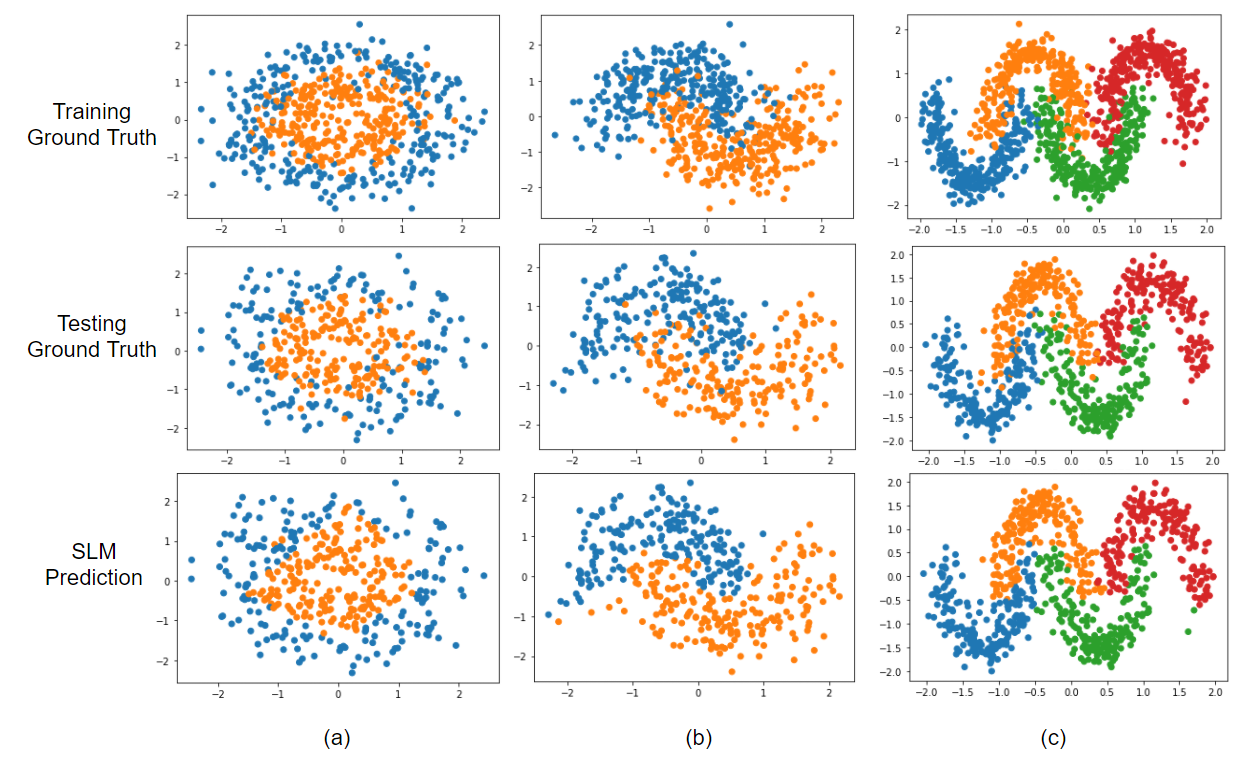}
\caption{Visualization of 2D feature datasets: (a) Circle \& Ring (b)
2-new-moon, (c) 4-new-moon. One ground truth sample of the training
data, the ground truth of the test data and the SLM predicted results
are shown in the first, second and third rows, respectively.} \label{fig:2}
\end{figure*}

{\em Datasets.} To evaluate the performance of SLM, we conduct
experiments on the following nine datasets. 
\begin{enumerate}
\item Circle-and-Ring. It contains an inner circle as one class and an
outer ring as the other class as shown in Fig.
\ref{fig:2}(a)~\cite{sklearn}. 
\item 2-New-Moons. It contains two interleaving new moons as shown in
Fig. \ref{fig:2}(b)~\cite{sklearn}. Each new moon corresponds to one
class. 
\item 4-New-Moons. It contains four interleaving new moons as shown in
Fig. \ref{fig:2}(c) \cite{sklearn}, where each moon is a class. 
\item Iris Dataset. The Iris plants dataset \cite{sklearn,Fisher1936}
has 3 classes, 4D features and 150 samples.
\item Wine Dataset. The Wine recognition dataset \cite{sklearn,UCI} has
3 classes, 13D features, and 178 samples.
\item B.C.W. Dataset. The breast cancer Wisconsin (B.C.W.) dataset
\cite{sklearn,UCI} has 2 classes, 30D features and 569 samples. 
\item Diabetes Dataset. The Pima Indians diabetes dataset \cite{diabete}
is for diabetes prediction. It has 2 classes, 8D features and 768
samples. By following \cite{ruiyuan}, we removed samples with the
physically impossible zero value for glucose, diastolic blood pressure,
triceps skin fold thickness, insulin, or BMI and used the remaining 392
samples for consistent experimental settings. 
\item Ionosphere Dataset. The Ionosphere binary classification dataset
\cite{ionosphere,UCI} is used to predict whether the radar return is
good or bad. It has 2 classes, 34D features and 351 instances. For
consistency with \cite{ruiyuan}, we remove the feature dimension that
has the zero variance from the data. 
\item Banknote Dataset. The banknote authentication dataset \cite{UCI}
classifies whether a banknote is genuine or forged based on the features
extracted from the wavelet transform of banknote images. It has 2 classes,
4D features and 1372 samples. 
\end{enumerate}
The feature dimension of the first three datasets is two while that of
the last six datasets is higher than two.  The first three are synthetic
ones, where 500 samples per class are generated with 30\% noisy samples
in the decision boundary for 2-New-Moons and 20\% noisy samples in the
decision boundary of Circle-and-Ring and 4-New-Moons.  The last six are
real-world datasets.  Samples in all datasets are randomly split into
training and test sets with 60\% and 40\%, respectively. 

{\em Benchmarking Classifiers and Their Implementations.} We compare the
performance of 10 classifiers in Table \ref{table:performance}.  They
include seven traditional classifiers and three proposed SLM variants.
The seven classifiers are: 1) MLP designed in a feedforward manner
(denoted by FF-MLP) \cite{ruiyuan}, 2) MLP trained by backpropagation
(denoted by BP-MLP), 3) linear SVM (LSVM), 4) SVM with thee radial basis
function (SVM/RBF) kernel, and 5) Decision Tree (DT) 6) Random Forest
(RF) and 7) XGBoost.  The three SLM variants are: 1) SLM Baseline using
only one SLM tree, 2) SLM Forest, and 3) SLM Boost. 

For the two MLP models, the network architectures of FF-MLP and BP-MLP
and their performance results are taken from \cite{ruiyuan}.  FF-MLP
has a four-layer network architecture; namely, one input layer, two
hidden layers and one output layer.  The neuron numbers of its input and
output layers are equal to the feature dimension and the class number,
respectively.  The neuron numbers at each hidden layer are
hyper-parameters determined adaptively by a dataset. BP-MLP has the same
architecture as FF-MLP against the same dataset.  Its model parameters
are initialized by those of FF-MLP and trained for 50 epochs.  For the
two SVM models, we conduct grid search for hyper-parameter $C$ in LSVM
and hyper-parameters $C$ and $\gamma$ in SVM/RBF for each of the nine
datasets to yield the optimal performance.  For the DT model, the
weighted entropy is used as the loss function in node splitting.  We do
not set the maximum depth limit of a tree, the minimum sample number and
the minimum loss decrease required as the stopping criteria. Instead, we
allow the DT for each dataset to split until the purest nodes are
reached in the training. For the ensemble of DT models (i.e., RF and
XGBoost), we conduct grid search for the optimal tree depth and the
learning rate of XGBoost to ensure that they reach the optimal
performance for each dataset. The number of trees is set to 100 to
ensure convergence.  The hyper-parameters of SLM Baseline (i.e., with one
SLM tree) include $D_{0}$, $p$, $A_{int}$, $\alpha$, $\beta$ and the
minimum number of samples used in the stopping criterion. They are
searched to achieve the performance as shown in \ref{table:performance}.
The number of trees in SLM Forest is set to 20 due to the faster
convergence of stronger individual SLM trees.  The number of trees in
SLM Boost is the same as that of XGBoost for fair comparison of
learning curves. 

{\em Comparison of Classification Performance.} The classification
accuracy results of 10 classifiers are shown in Table
\ref{table:performance}. We divide the 10 classifiers into two groups.
The first group includes 6 basic methods: FF-MLP, BP-MLP, LSVM, SVM/RBF,
DT and SLM Baseline. The second group includes 4 ensemble methods: RF,
XGBoost, SLM Forest and SLM Boost.  The best results for each group
are shown in bold.  For the first group, SLM Baseline and SVM/RBF often
outperform FF-MLP, BP-MLP, LSVM and DT and give the best results.  For
the three synthetic 2D datasets (i.e.  circle-and-ring, 2-new-moons and
4-new-moons), the gain of SLM over MLP is relatively small due to noisy
samples. The difference in sample distributions of training and test
data plays a role in the performance upper bound.  To demonstrate this
point, we show sample distributions of their training and testing data
in Fig.  \ref{fig:2}.  For the datasets with high dimensional input
features, the SLM methods achieve better performance over the classical
methods with bigger margins.  The ensemble methods in the second group
have better performance than the basic methods in the first group. The
ensembles of SLM are generally more powerful than those of DT. They give
the best performance among all benchmarking methods. 

\begin{table*}[!htbp]
\begin{center}
\caption{Model size comparison of five machine learning models models
against 9 datasets, where the smallest model size for each dataset is 
highlighted in bold.}\label{tab3}
\begin{tabular}{|l|c|c|c|c|c||c|c|} \hline
      Datasets & FF/BP-MLP & LSVM & SVM/RBF & DT  & SLM Baseline & DT depth & SLM tree depth\\ \hline
      Circle-and-Ring & 125	& 2,965&	1,425&	350	& \textbf{39} &14 &	4\\
      2-new-moons & 114	&1,453&	1,305	&286	&\textbf{42} &	15	&4\\
      4-new-moons &702&	2,853&	2,869&	298	&\textbf{93} &	11&	5\\
      Iris & 47	& 235&	343 &	34		&\textbf{20} &6	&3\\
      Wine & 147&	453	&  963	& \textbf{26}	&99  &4 &	2  \\
      B.C.W. &74&	1,462&	3,254 &	\textbf{54}	&126  &7 &	4 \\
      Pima & 2,012& 1,532 & 1,802 &	130	&55  &	8	&3\\
      Ionosphere & 278	&1,017 &	2,207 &	\textbf{50}	&78	  &10 &2\\
      Banknode & 22	& 1,160 & 1,322 & 78&	\textbf{40} 	&7 &	3 \\\hline
\end{tabular}
\end{center}
\end{table*}

{\em Comparison of Model Sizes.} The model size is defined by the number
of model parameters.  The model sizes of FF/BP-MLP, LSVM, SVM/RBF, DT
and SLM Baseline are compared in Table \ref{tab3}.  Since FF-MLP and
BP-MLP share the same architecture, their model sizes are the same. It
is calculated by summing up the weight and bias numbers of all neurons. 

The model parameters of LSVM and SVM/RBF can be computed as
\begin{equation}\label{eqn:SVM_size}
\mbox{SVM Parameter \#} = L + 1 + (D + 2)N_{SV},
\end{equation} 
where $L$, $D$ and $N_{SV}$ denote the number of training samples, the
feature dimension and the number of support vectors, respectively.  The
first term in Eq. (\ref{eqn:SVM_size}) is the slack variable for each
training sample. The second term denotes the bias. The last term comes
from the fact that each support vector has $D$ feature dimensions, one
Lagrange dual coefficient, and one class label. 

The model sizes of DTs depend on the number of splits learned during the
training process, and there are two parameters learned during each split
for feature selection and split value respectively, the sizes of DTs are
calculated as two times of the number of splits. 

The size of an SLM baseline model depends on the number of partitioning
hyper-planes which are determined by the training stage. For given
hyper-parameter $\bm{D}_{0}$, each partitioning hyper-plane involves one
weight matrix and a selected splitting threshold, with $\bm{q}_{i}$
decorrelated partitioning learned for partitioning each parent node.
Then, the model size of the corresponding SLM can be calculated as
\begin{eqnarray}
\mbox{SLM Parameter \#} = \sum^{M}_{i=1} \bm{q}_{i} (\bm{D}_{0} + 1),
\end{eqnarray} 
where $M$ is the number of partitioning hyperplanes.

Details on the model and computation of each method against each dataset
are given in the appendix.  It is worthwhile to comment on the tradeoff
between the classification performance and the model size. For SVM,
since its training involves learning the dual coefficients and slack
variables for each training sample and memorization of support vectors,
the model size is increasing linearly with the number of training
samples and the number of support vectors. When achieving similar
classification accuracy, the SVM model is heavier for more training
samples.  For MLPs, with high-dimension classification tasks, SLM
methods outperforms the MLP models in all benchmarking datasets. For the
datasets with saturated performance such as Iris, Banknote, and
Ionosphere, SLM achievs better or comparable performance with less than
half of the parameters of MLP.  As compared to DTs, the SLM models tend
to achieve wider and shallower trees, as shown in Table \ref{tab3}. The
depth of SLM trees are overall smaller than the DT models, while the
number of splittings can be comparable for the small datasets, like
Wine. The SLM trees tend to make more splits at reach pure leaf nodes at
shallower depth. With outperforming the DTs in all the datasets, the SLM
model sizes are generally smaller than the DTs as well with benefiting
from the subspace partitioning process. 

\begin{figure}[t]
\begin{center}
\includegraphics[width=1\linewidth]{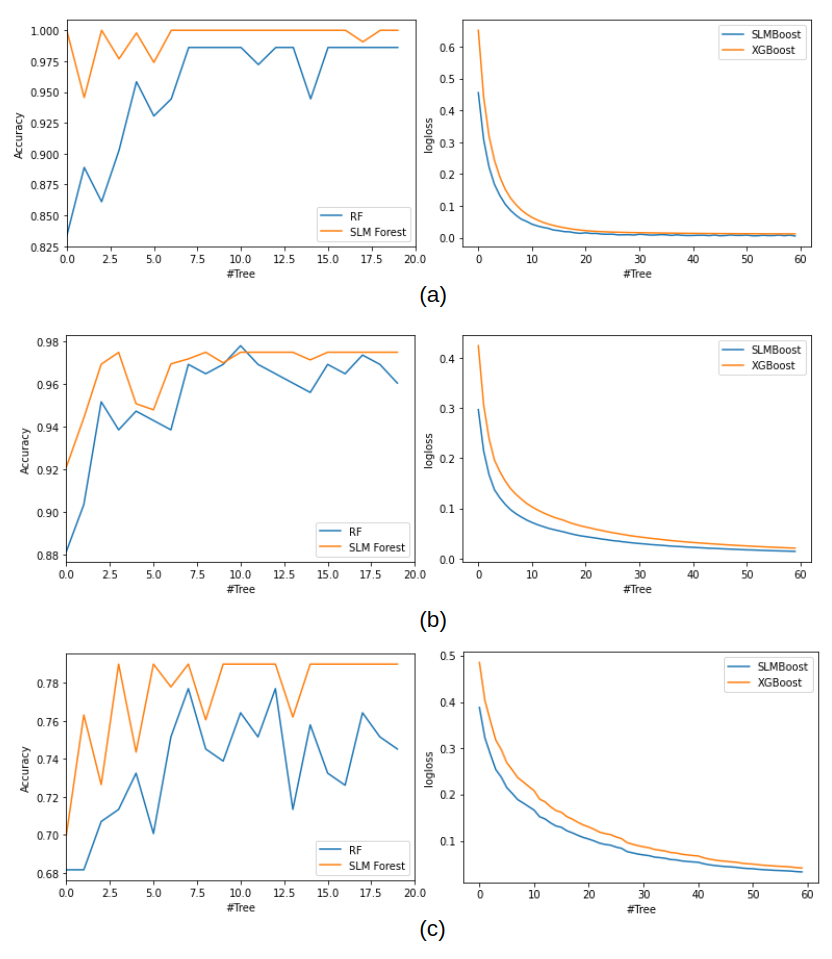} 
\end{center}
\caption{Comparison of SLM and DT ensembles for three datasets (a) Wine,
(b) B.C.W., and (c) Pima. Each left subfigure compares the accuracy
curves of SLM Forest and RF as a function of the tree number. Each right
subfigure compares the logloss curves of SLM Boost and XGBoost as a
function of the tree number.}\label{fig:3}
\end{figure}

\begin{table*}[!htbp]
\begin{center}
\caption{Comparison of regression performance of eight regressors on six datasets.}\label{table:SLR} 
    \begin{tabular}{|l|c|c|c|c|c|c|} \hline
      \textbf{Datasets} & Make Friedman1&	Make Friedman2&	Make Friedman3&	Boston&	California\textunderscore housing & Diabetes \\ \hline
        LSVR 	&2.49	&138.43	&0.22	&4.90	&0.76 &53.78\\
        SVR/RBF	&\textbf{1.17}	&\textbf{6.74}	&\textbf{0.11}&\textbf{3.28}	&\textbf{0.58} &\textbf{53.71}\\
        DT	&3.10	&33.57	&0.11	&4.75	&0.74 &76.56\\
        SLR Baseline&2.89	&31.28	&0.11	&4.42	&0.69 &56.05\\ \hline \hline
        RF	&2.01	&22.32	&0.08	&3.24	&0.52 &54.34\\
        XGBoost	&1.17	&32.34	&0.07	&2.67	&0.48 &53.99\\
        SLR Forest&1.88	&20.79	&0.08	&3.01	&0.48 &52.52\\
        SLR Boost &\textbf{1.07} &\textbf{18.07}&\textbf{0.06}&\textbf{2.39}&\textbf{0.45}&\textbf{51.27}\\\hline
    \end{tabular}
\end{center}
\end{table*}

{\em Convergence Performance Comparison of DT Ensembles and SLM
Ensembles.} We compare the convergence performance of the ensemble and
the boosting methods of DT and SLM for Wine, B.C.W. and Pima three datas
in Figs. \ref{fig:3}(a)-(c). For RF and SLM Forest, which are ensembles
of DT and SLM, respectively, we set their maximum tree depth and
learning rate to the same. We show their accuracy curves as a function
of the tree number in the left subfigure. We see that SLM Forest
converges faster than RF. For XGBoost and SLM Boost, which are boosting
methods of DT and SLM, respectively, we show the logloss value as a
function of the tree number in the right subfigure. Again, we see that
SLM Boost converges faster than XGBoost. 


\section{Subspace Learning Regressor (SLR)}\label{sec:SLR}

\subsection{Method}\label{subsec:SLR_algorithm}

A different loss function can be adopted in the subspace partitioning
process for a different task. For example, to solve a regression
problem, we can follow the same methodology as described in Sec.
\ref{sec:method} but adopt the mean-squrared-error (MSE) as the loss
function.  The resulting method is called subspace learning regression,
and the corresponding regressor is the subspace learning regressor
(SLR).

Mathematically, each training sample has a pair of input ${\bf x}$ and
output $y$, where ${\bf x}$ is a $D$-dimensional feature vector and $y$
is a scalar that denotes the the regression target. Then, we build an
SLR tree that partitions the $D$-dimensional feature space
hierarchically into a set of leaf nodes. Each of them corresponds to a
subspace.  The mean of sample targets in a leaf node is set as the
predicted regression value of these samples.  The partition objective is to reduce the total MSE of sample targets as much as possible.  In the
partitioning process, the total MSE of all leaf nodes decreases
gradually and saturates at a certain level. 

The ensemble and boosting methods are applicable to SLR. The SLR Forest
consists of multiple SLR trees through ensembles. Its final prediction
is the mean of predictions from SLR trees in the forest.  To derive SLR
Boost, we apply the GBDT process and train a series of additive SLR trees
to achieve gradient boosting, leading to further performance
improvement.  As compared with a decision tree, an SLR tree is wider,
shallower and more effective. As a result, SLR Forest and SLR Boost
are more powerful than their counter parts as demonstrated in the
next subsection.

\subsection{Performance Evaluation}\label{subsec:SLR_experiments}

To evaluate the performance of SLR, we compare the root mean squared
error (RMSE) performance of eight regressors on six datasets in Table
\ref{table:SLR}. The five benchmarking regressors are: linear SVR
(LSVR), SVR with RBF kernel, DT, RF and XGBoost. There are three
variants of SLR: SLR Baseline (with one SLR tree), SLR Forest and SLR
Boost. The first three datasets are synthesized datasets as described in
\cite{friedman}. We generate 1000 samples for each of them.  The last
three datasets are real world datasets.  Samples in all six datasets are
randomly split into 60\% training samples and 40\% test samples. 

\begin{enumerate} 
\item {\bf Make Friedman 1}. Its input vector, ${\bf x}$, contains $P$ (with
$P > 5$) elements, which are independent and uniformly distributed on
interval $[0, 1]$. Its output, $y$, is generated by the first five
elements of the input. The remaining $(P-5)$ elements are irrelevant
features and can be treated as noise. We refer to \cite{friedman} for
details.  We choose $P=10$ in the experiment. 
\item {\bf Make Friedman 2-3}.  Their input vector, ${\bf x}$, has 4
elements.  They are independent and uniformly distributed on interval
$[0, 1]$. Their associated output, $y$, can be calculated by all four
input elements via mathematical formulas as described in
\cite{friedman}. 
\item {\bf Boston}. It contains 506 samples, each of which has a 13-D
feature vector as the input. An element of the feature vector is a real
positive number. Its output is a real number within interval $[5,50]$. 
\item {\bf California Housing}. It contains 20640 samples, each of which
has an 8-D feature vector. The regression target (or output) is a real
number within interval $[0.15,5]$. 
\item {\bf Diabetes}. It contains 442 samples, each of which has a 10-D
feature vector.  Its regression target is a real number within interval
$[25, 346]$. 
\end{enumerate}

As shown in Table \ref{table:SLR}, SLR Baseline outperforms DT in all
datasets. Also, SLR Forest and SLR Boosting outperform RF and XGBoost,
respectively. For Make Friedman1, Make Friedman3, california-housing,
Boston, and diabetes, SLR Boost achieves the best performance. For Make Friedman2, SVR/RBF achieves the best performance benefiting from the
RBF on its specific data distribution.  However, it is worthwhile to
emphasize that, to achieve the optimal performance, SVR/RBF needs to
overfit to the training data by finetuning the soft margin with a large
regularization parameter (i.e., $C=1000$). This leads to much higher
computational complexity. With stronger individual SLR trees and
effective uncorrelated models, the ensemble of SLR can achieve better
performance than DTs with efficiency. 

\section{Comments on Related Work} \label{sec:review}

In this section, related prior work is reviewed and the relation between
various classifers/regressors and SLM/SLR are commented. 

\subsection{Classification and Regression Tree (CART)}\label{subsec:CDT}

DT has been well studied for decades, and is broadly applied for general
classification and regression problems. Classical decision tree
algorithms, e.g. ID3 \cite{ID3} and CART \cite{CART}, are devised to
learn a sequence of binary decisions.  One tree is typically a weak
classifier, and multiple trees are built to achieve higher performance
in practice such as bootstrap aggregation \cite{RF} and boosting methods
\cite{xgb}. DT and its ensembles work well most of the time. Yet, they
may fail due to poor training and test data splitting and training data
overfit. As compared to classic DT, one SLM tree (i.e., SLM Baseline)
can exploit discriminant features obtained by probabilistic projections
and achieve multiple splits at one node. SLM generally yields wider and
shallower trees. 

\subsection{Random Forest (RF)}\label{subsec:RF}

RF consists of multiple decisions trees. It is proved in
\cite{amit1997shape} that the predictive performance of RF depends on
the strength of individual trees and a measure of their dependence.  For
the latter, the lower the better. To achieve higher diversity, RF
training takes only a fraction of training samples and features in
building a tree, which trades the strength of each DT for the general
ensemble performance. In practice, several effective designs are proposed
to achieve uncorrelated individual trees. For example, bagging
\cite{bagging} builds each tree through random selection with
replacement in the training set. Random split selection
\cite{randomsplit} selects a split at a node among the best splits at
random. In \cite{randomsubspace}, a random subset of features is
selected to grow each tree. Generally speaking, RF uses bagging and
feature randomness to create uncorrelated trees in a forest, and their
combined prediction is more accurate than that of an individual tree.
In contrast, the tree building process in SLM Forest always takes all
training samples and the whole feature space into account. Yet, it
utilizes feature randomness to achieve the diversity of each SLM tree as
described in Sec. \ref{subsec:forest}. Besides effective diversity of
SLM trees, the strength of each SLM tree is not affected in SLM Forest.
With stronger individual learners and effective diversity, SLM Forest
achieves better predictive performance and faster convergence in terms
of the tree number. 

\subsection{Gradient Boosting Decision Tree (GBDT)}\label{subsec:GBDT}

Gradient boosting is another ensemble method of weak learners. It
builds a sequence of weak prediction models. Each new model attempts
to compensate the prediction residual left in previous models.  The
gradient boosting decision tree (GBDT) methods includes \cite{GBDT},
which performs the standard gradient boosting, and XGBoost
\cite{chen2015xgboost, chen2016xgboost}, which takes the Taylor series
expansion of a general loss function and defines a gain so as to perform
more effective node splittings than standard DTs.  SLM Boost mimics the
boosting process of XGBoost but replaces DTs with SLM trees.  As
compared with standard GBDT methods, SLM Boost achieves faster
convergence and better performance as a consequence of stronger
performance of an SLM tree. 

\subsection{Multilayer Perceptron (MLP)}\label{subsec:MLP}

Since its introduction in 1958 \cite{rosenblatt1958perceptron}, MLP has
been well studied and broadly applied to many classification and
regression tasks \cite{speech,economic, ImageProcessing}. Its universal
approximation capability is studied in \cite{cybenko1989approximation,
HornikEtAl89, Stinchcombe1989UniversalAU,
Leshno93multilayerfeedforward}.  The design of a practical MLP solution
can be categorized into two approaches.  First, one can propose an
architecture and fine tune parameters at each layer through back
propagation. For the MLP architecture, it is often to evaluate different
networks through trials and errors. Design strategies include tabu
search \cite{tabu} and simulated annealing \cite{annealing}. There are
quite a few MLP variants.  In convolutional neural networks (CNNs)
\cite{lecun1998gradient, lenet5, alexnet, deeplearning}, convolutional
layers share neurons' weights and biases across different spatial
locations while fully-connected layers are the same as traditional MLPs.
MLP also serves as the building blocks in transformer models
\cite{transformer,vit}.  Second, one can design an MLP by constructing
its model layer after layer, e.g., \cite{parekh1997constructive,
mezard1989learning, frean1990upstart, parekh2000constructive}. To
incrementally add new layers, some suggest an optimization method which
is similar to the first approach, e.g.  \cite{Kwok1997-sq,
gallant1990perceptron, mascioli1995constructive}. They adjust the
parameters of the newly added hidden layer and determine the parameters
without back propagation, e.g.  \cite{geo, yang1999distal,
marchand1989learning}.

The convolution operation in neural networks changes the input feature
space to an output feature space, which serves as the input to the next
layer. Nonlinear activation in a neuron serves as a partition of the
output feature space and only one half subspace is selected to resolve
the sign confusion problem caused by the cascade of convolution
operations \cite{kuo2016understanding, ruiyuan}.  In contrast, SLM does
not change the feature space at all. The probabilistic projection in SLM
is simply used for feature space partitioning without generating new
features.  Each tree node splitting corresponds to a hyperplane
partitioning through weights and bias learning. As a result, both half
subspaces can be preserved. SLM learns partitioning parameters in a
feedforward and probabilistic approach, which is efficient and
transparent. 

\subsection{Extreme Learning Machine (ELM)}\label{subsec:ELM}

ELM \cite{ELM} adopts random weights for the training of feedforward
neural networks. Theory of random projection learning models and their
properties (e.g., interpolation and universal approximation) have been
investigated in \cite{huang2006extreme, huang2006universal,
huang2007convex, huang2008enhanced}. To build MLP with ELM, one can add
new layers with randomly generated weights. However, the long training
time and the large model size due to a large search space imposes their
constraints in practical applications. SLM Baseline does take the
efficiency into account. It builds a general decision tree through
probabilistic projections, which reduces the search space by leveraging
most discriminant features with several hyper-parameters. We use the
term ``probabilistic projection" rather than ``random projection" to
emphasize their difference.

\section{Conclusion and Future Work}\label{sec:conclusion}

In this paper, we proposed a novel machine learning model called the
subspace learning machine (SLM).  SLM combines feedforward multilayer
perceptron design and decision tree to learn discriminate subspace and
make predictions. At each subspace learning step, SLM utilizes
hyperplane partitioning to evaluate each feature dimension, and probabilistic projection to learn parameters for perceptrons for feature learning, the
most discriminant subspace is learned to partition the data into child
SLM nodes, and the subspace learning process is conducted iteratively,
final predictions are made with pure SLM child nodes.  SLM is
light-weight, mathematically transparent, adaptive to high dimensional
data, and achieves state-of-the-art benchmarking performance. SLM tree
can serve as weak classifier in general boosted and bootstrap
aggregation methods as a more generalized model. 

Recently, research on unsupervised representation learning for images
has been intensively studied, e.g., \cite{chen2018saak,
chen2020pixelhop, chen2020pixelhop++, kuo2018data,
kuo2019interpretable}. The features associated with their labels can be
fed into a classifier for supervised classification. Such a system
adopts the traditional pattern recognition paradigm with feature
extraction and classification modules in cascade.  The feature dimension
for these problems are in the order of hundreds.  It will be interesting
to investigate the effectiveness of SLM for problems of the
high-dimensional input feature space. 

\section*{Appendix}

The sizes of the classification models in Table 2 are computed below.

\begin{itemize}
\item Circle-and-Ring. \\
FF-MLP has four Gaussian components for the ring and one Gaussian blob
for the circle.  As a result, there are 8 and 9 neurons in the two
hidden layers with 125 parameters in total.  LSVM has 600 slack
variables, 591 support vectors and one bias, resulting in 2,965
parameters.  SVM/RBF kernel has 600 slack variables, 206 support vectors
and one bias, resulting in 1,425 parameters.  DT has 175 splits to
generate a tree of depth 14, resulting in 350 parameters.  SLM utilizes
two input features to build a tree of depth 4. The node numbers at each
level are 1, 4, 4, 10 and 8, respectively. It has 13 partitions and 39
parameters. 

\item 2-New-Moons. \\
FF-MLP has 2 Gaussian components for each class, 8 neurons in each of
two hidden layers, and 114 parameters in total.  LSVM has 600 slack
variables, 213 support vectors and one bias, resulting in 1,453
parameters.  SVM/RBF kernel has 600 slack variables, 176 support vectors
and one bias, resulting in 1,305 parameters.  DT has 143 splits to
generate a tree of depth 15, resulting in 286 parameters.  SLM has a
tree of depth 4 and the node numbers at each level are 1, 4, 8, 12 and
4, respectively. It has 14 partitions and 42 parameters. 

\item 4-New-Moons. \\
FF-MLP has three Gaussian components for each class, 18 and 28 neurons
in two hidden layers, respectively, and 702 parameters in total.  LSVM
has 1200 slack variables, 413 support vectors and one bias, resulting in
2,853 parameters.  SVM/RBF has 1200 slack variables, 417 support vectors
and one bias, resulting in 2,869 parameters.  DT for this dataset made
149 splits with max depth of the tree as 11, results in 298 parameters.
SLM has a tree is of depth 5 and the node numbers at each level are 1,
4, 16, 22, 16 and 4, respectively. It has 31 partitions and 93
parameters. 

\item Iris Dataset. \\
FF-MLP has two Gaussian components for each class and two partitioning
hyper-planes. It has 4 and 3 neurons at hidden layers 1 and 2,
respectively, and 47 parameters in total.  LSVM has 90 slack variables,
24 support vectors and one bias, resulting in 235 parameters.  SVM/RBF
has 90 slack variables, 42 support vectors and one bias, resulting in
343 parameters.  DT has 17 splits with a tree of max depth 6, results in
34 parameters.  SLM uses all 4 input dimensions and has a tree of depth
3. The node numbers at each level are 1, 2, 2 and 4, respectively.  It
has 4 partitions and 20 parameters. 

\item Wine Dataset. \\
FF-MLP assigns two Gaussian components to each class. There are 6
neurons at each of two hidden layers. It has 147 parameters.  LSVM has
107 slack variables, 23 support vectors and one bias, resulting in 453
parameters.  SVM/RBF has 107 slack variables, 57 support vectors and one
bias, resulting in 963 parameters.  DT has 13 splits with a tree of max
depth 4, results in 26 parameters.  SLM sets $\bm{D}_{0}= 8$ and has a
tree of depth 2. There are 1, 8 and 256 nodes at each level,
respectively. It has 11 partitions and 99 parameters. 

\item B.C.W. Dataset. \\
FF-MLP assigns two Gaussian components to each class. There are two
neurons at each of the two hidden layers. The model has 74 parameters.
LSVM has 341 slack variables, 35 support vectors and one bias, resulting
in 1,462 parameters.  SVM with RBF kernel has 341 slack variables, 91
support vectors and 1 bias, resulting in 3,254 parameters.  DT has 27
splits with a tree of max depth 7, results in 54 parameters.  SLM sets
$\bm{D}_{0} = 5$ and has a tree of depth 4 with 1, 8, 16, 8 and 32 nodes
at each level, respectively. It has 21 partitions and 126 parameters. 
 
\item Diabetes Dataset. \\
FF-MLP has 18 and 88 neurons at two hidden layers, respectively. The
model has 2,012 parameters.  LSVM has 461 slack variables, 107 support
vectors and one bias, resulting in 1,532 parameters.  SVM/RBF has 461
slack variables, 134 support vectors and one bias, resulting in 1,802
parameters.  DT has 65 splits with a tree of max depth 8, results in 130
parameters.  SLM sets $\bm{D}_{0}=4$ and has a tree of depth 3 with 1,
2, 16 and 20 nodes at each level, respectively.  It has 11 partitions
and 55 parameters. 

\item Ionosphere Dataset. \\
FF-MLP has 6 and 8 neurons in hidden layers 1 and 2, respectively. It
has 278 parameters.  LSVM has 211 slack variables, 23 support vectors
and one bias, resulting in 1017 parameters.  SVM/RBF has 211 slack
variables, 57 support vectors and one bias, resulting in 2,207
parameters.  DT has 25 splits with a tree of max depth 10, resulting in
50 parameters.  SLM sets $\bm{D}_{0} = 5$ and has a tree of depth 2 with
1, 4 and 20 nodes at each level, respectively.  It has 13 partitions and
78 parameters. 

\item Banknote Dataset. \\
FF-MLP has two neurons at each of the two hidden layers. It has 22
parameters.  LSVM has 823 slack variables, 56 support vectors and one
bias, resulting in 1,160 parameters.  SVM/RBF has 823 slack variables,
83 support vectors and one bias, resulting in 1,322 parameters.  DT has
39 splits with a tree of max depth 7, resulting in 78 parameters.  SLM
uses all input features and has a tree of depth 3 with 1, 2, 8 and 16
nodes at each level, respectively.  It has 8 partitions and 40
parameters. 

\end{itemize}

\bibliographystyle{IEEEtran}
\bibliography{refs}

\begin{thebibliography}{10}
\providecommand{\url}[1]{#1}
\csname url@samestyle\endcsname
\providecommand{\newblock}{\relax}
\providecommand{\bibinfo}[2]{#2}
\providecommand{\BIBentrySTDinterwordspacing}{\spaceskip=0pt\relax}
\providecommand{\BIBentryALTinterwordstretchfactor}{4}
\providecommand{\BIBentryALTinterwordspacing}{\spaceskip=\fontdimen2\font plus
\BIBentryALTinterwordstretchfactor\fontdimen3\font minus
  \fontdimen4\font\relax}
\providecommand{\BIBforeignlanguage}[2]{{%
\expandafter\ifx\csname l@#1\endcsname\relax
\typeout{** WARNING: IEEEtran.bst: No hyphenation pattern has been}%
\typeout{** loaded for the language `#1'. Using the pattern for}%
\typeout{** the default language instead.}%
\else
\language=\csname l@#1\endcsname
\fi
#2}}
\providecommand{\BIBdecl}{\relax}
\BIBdecl

\bibitem{svm}
C.~Cortes and V.~Vapnik, ``Support-vector networks,'' \emph{Machine learning},
  vol.~20, no.~3, pp. 273--297, 1995.

\bibitem{CART}
L.~Breiman, J.~Friedman, C.~Stone, and R.~Olshen, ``Classification and
  regression trees (crc, boca raton, fl),'' 1984.

\bibitem{rosenblatt1958perceptron}
F.~Rosenblatt, ``The perceptron: a probabilistic model for information storage
  and organization in the brain.'' \emph{Psychological review}, vol.~65, no.~6,
  p. 386, 1958.

\bibitem{ruiyuan}
R.~Lin, Z.~Zhou, S.~You, R.~Rao, and C.-C.~J. Kuo, ``From two-class linear
  discriminant analysis to interpretable multilayer perceptron design,''
  \emph{arXiv preprint arXiv:2009.04442}, 2020.

\bibitem{ELM}
G.-B. Huang, Q.-Y. Zhu, and C.-K. Siew, ``Extreme learning machine: theory and
  applications,'' \emph{Neurocomputing}, vol.~70, no. 1-3, pp. 489--501, 2006.

\bibitem{RF}
L.~Breiman, ``Random forests,'' \emph{Machine learning}, vol.~45, no.~1, pp.
  5--32, 2001.

\bibitem{GBDT}
J.~H. Friedman, ``Greedy function approximation: a gradient boosting machine,''
  \emph{Annals of statistics}, pp. 1189--1232, 2001.

\bibitem{yang2022supervised}
Y.~Yang, W.~Wang, H.~Fu, and C.-C.~J. Kuo, ``On supervised feature selection
  from high dimensional feature spaces,'' \emph{arXiv preprint
  arXiv:2203.11924}, 2022.

\bibitem{chen2015xgboost}
T.~Chen, T.~He, M.~Benesty, V.~Khotilovich, Y.~Tang, H.~Cho, K.~Chen
  \emph{et~al.}, ``Xgboost: extreme gradient boosting,'' \emph{R package
  version 0.4-2}, vol.~1, no.~4, pp. 1--4, 2015.

\bibitem{chen2016xgboost}
T.~Chen and C.~Guestrin, ``Xgboost: A scalable tree boosting system,'' in
  \emph{Proceedings of the 22nd acm sigkdd international conference on
  knowledge discovery and data mining}, 2016, pp. 785--794.

\bibitem{amit1997shape}
Y.~Amit and D.~Geman, ``Shape quantization and recognition with randomized
  trees,'' \emph{Neural computation}, vol.~9, no.~7, pp. 1545--1588, 1997.

\bibitem{sklearn}
F.~Pedregosa, G.~Varoquaux, A.~Gramfort, V.~Michel, B.~Thirion, O.~Grisel,
  M.~Blondel, P.~Prettenhofer, R.~Weiss, V.~Dubourg \emph{et~al.},
  ``Scikit-learn: Machine learning in python,'' \emph{the Journal of machine
  Learning research}, vol.~12, pp. 2825--2830, 2011.

\bibitem{Fisher1936}
R.~A. Fisher, ``The use of multiple measurements in taxonomic problems,''
  \emph{Annals of eugenics}, vol.~7, no.~2, pp. 179--188, 1936.

\bibitem{UCI}
A.~Asuncion and D.~Newman, ``Uci machine learning repository,'' 2007.

\bibitem{diabete}
J.~W. Smith, J.~E. Everhart, W.~Dickson, W.~C. Knowler, and R.~S. Johannes,
  ``Using the adap learning algorithm to forecast the onset of diabetes
  mellitus,'' in \emph{Proceedings of the annual symposium on computer
  application in medical care}.\hskip 1em plus 0.5em minus 0.4em\relax American
  Medical Informatics Association, 1988, p. 261.

\bibitem{ionosphere}
J.~P. G{\"o}pfert, H.~Wersing, and B.~Hammer, ``Interpretable locally adaptive
  nearest neighbors,'' \emph{Neurocomputing}, vol. 470, pp. 344--351, 2022.

\bibitem{friedman}
J.~H. Friedman, ``Multivariate adaptive regression splines,'' \emph{The annals
  of statistics}, vol.~19, no.~1, pp. 1--67, 1991.

\bibitem{ID3}
J.~R. Quinlan, ``Induction of decision trees,'' \emph{Machine learning},
  vol.~1, no.~1, pp. 81--106, 1986.

\bibitem{xgb}
T.~Chen and C.~Guestrin, ``Xgboost: A scalable tree boosting system,'' in
  \emph{Proceedings of the 22nd acm sigkdd international conference on
  knowledge discovery and data mining}, 2016, pp. 785--794.

\bibitem{bagging}
L.~Breiman, ``Bagging predictors,'' \emph{Machine learning}, vol.~24, no.~2,
  pp. 123--140, 1996.

\bibitem{randomsplit}
T.~G. Dietterich, ``An experimental comparison of three methods for
  constructing ensembles of decision trees: Bagging, boosting, and
  randomization,'' \emph{Machine learning}, vol.~40, no.~2, pp. 139--157, 2000.

\bibitem{randomsubspace}
T.~K. Ho, ``The random subspace method for constructing decision forests,''
  \emph{IEEE transactions on pattern analysis and machine intelligence},
  vol.~20, no.~8, pp. 832--844, 1998.

\bibitem{speech}
A.~Ahad, A.~Fayyaz, and T.~Mehmood, ``Speech recognition using multilayer
  perceptron,'' in \emph{IEEE Students Conference, ISCON'02. Proceedings.},
  vol.~1.\hskip 1em plus 0.5em minus 0.4em\relax IEEE, 2002, pp. 103--109.

\bibitem{economic}
A.~V. Devadoss and T.~A.~A. Ligori, ``Forecasting of stock prices using multi
  layer perceptron,'' \emph{International journal of computing algorithm},
  vol.~2, no.~1, pp. 440--449, 2013.

\bibitem{ImageProcessing}
K.~Sivakumar and U.~B. Desai, ``Image restoration using a multilayer perceptron
  with a multilevel sigmoidal function,'' \emph{IEEE transactions on signal
  processing}, vol.~41, no.~5, pp. 2018--2022, 1993.

\bibitem{cybenko1989approximation}
G.~Cybenko, ``Approximation by superpositions of a sigmoidal function,''
  \emph{Mathematics of control, signals and systems}, vol.~2, no.~4, pp.
  303--314, 1989.

\bibitem{HornikEtAl89}
K.~Hornik, M.~Stinchcombe, and H.~White, ``Multilayer feedforward networks are
  universal approximators,'' \emph{Neural networks}, vol.~2, no.~5, pp.
  359--366, 1989.

\bibitem{Stinchcombe1989UniversalAU}
M.~Stinchombe, ``Universal approximation using feed-forward networks with
  nonsigmoid hidden layer activation functions,'' \emph{Proc. IJCNN,
  Washington, DC, 1989}, pp. 161--166, 1989.

\bibitem{Leshno93multilayerfeedforward}
M.~Leshno, V.~Y. Lin, A.~Pinkus, and S.~Schocken, ``Multilayer feedforward
  networks with a nonpolynomial activation function can approximate any
  function,'' \emph{Neural networks}, vol.~6, no.~6, pp. 861--867, 1993.

\bibitem{tabu}
F.~Glover, ``Future paths for integer programming and links to artificial
  intelligence,'' \emph{Computers \& operations research}, vol.~13, no.~5, pp.
  533--549, 1986.

\bibitem{annealing}
S.~Kirkpatrick, C.~D. Gelatt~Jr, and M.~P. Vecchi, ``Optimization by simulated
  annealing,'' \emph{science}, vol. 220, no. 4598, pp. 671--680, 1983.

\bibitem{lecun1998gradient}
Y.~LeCun, L.~Bottou, Y.~Bengio, and P.~Haffner, ``Gradient-based learning
  applied to document recognition,'' \emph{Proceedings of the IEEE}, vol.~86,
  no.~11, pp. 2278--2324, 1998.

\bibitem{lenet5}
Y.~LeCun, B.~Boser, J.~S. Denker, D.~Henderson, R.~E. Howard, W.~Hubbard, and
  L.~D. Jackel, ``Backpropagation applied to handwritten zip code
  recognition,'' \emph{Neural computation}, vol.~1, no.~4, pp. 541--551, 1989.

\bibitem{alexnet}
A.~Krizhevsky, I.~Sutskever, and G.~E. Hinton, ``Imagenet classification with
  deep convolutional neural networks,'' \emph{Advances in neural information
  processing systems}, vol.~25, 2012.

\bibitem{deeplearning}
Y.~LeCun, Y.~Bengio, and G.~Hinton, ``Deep learning,'' \emph{nature}, vol. 521,
  no. 7553, pp. 436--444, 2015.

\bibitem{transformer}
A.~Vaswani, N.~Shazeer, N.~Parmar, J.~Uszkoreit, L.~Jones, A.~N. Gomez,
  {\L}.~Kaiser, and I.~Polosukhin, ``Attention is all you need,''
  \emph{Advances in neural information processing systems}, vol.~30, 2017.

\bibitem{vit}
A.~Dosovitskiy, L.~Beyer, A.~Kolesnikov, D.~Weissenborn, X.~Zhai,
  T.~Unterthiner, M.~Dehghani, M.~Minderer, G.~Heigold, S.~Gelly \emph{et~al.},
  ``An image is worth 16x16 words: Transformers for image recognition at
  scale,'' \emph{arXiv preprint arXiv:2010.11929}, 2020.

\bibitem{parekh1997constructive}
R.~Parekh, J.~Yang, and V.~Honavar, ``Constructive neural network learning
  algorithms for multi-category real-valued pattern classification,''
  \emph{Dept. Comput. Sci., Iowa State Univ., Tech. Rep. ISU-CS-TR97-06}, 1997.

\bibitem{mezard1989learning}
M.~M{\'e}zard and J.-P. Nadal, ``Learning in feedforward layered networks: The
  tiling algorithm,'' \emph{Journal of Physics A: Mathematical and General},
  vol.~22, no.~12, p. 2191, 1989.

\bibitem{frean1990upstart}
M.~Frean, ``The upstart algorithm: A method for constructing and training
  feedforward neural networks,'' \emph{Neural computation}, vol.~2, no.~2, pp.
  198--209, 1990.

\bibitem{parekh2000constructive}
R.~Parekh, J.~Yang, and V.~Honavar, ``Constructive neural-network learning
  algorithms for pattern classification,'' \emph{IEEE Transactions on neural
  networks}, vol.~11, no.~2, pp. 436--451, 2000.

\bibitem{Kwok1997-sq}
T.-Y. Kwok and D.-Y. Yeung, ``Objective functions for training new hidden units
  in constructive neural networks,'' \emph{IEEE Transactions on neural
  networks}, vol.~8, no.~5, pp. 1131--1148, 1997.

\bibitem{gallant1990perceptron}
S.~I. Gallant \emph{et~al.}, ``Perceptron-based learning algorithms,''
  \emph{IEEE Transactions on neural networks}, vol.~1, no.~2, pp. 179--191,
  1990.

\bibitem{mascioli1995constructive}
F.~F. Mascioli and G.~Martinelli, ``A constructive algorithm for binary neural
  networks: The oil-spot algorithm,'' \emph{IEEE Transactions on Neural
  Networks}, vol.~6, no.~3, pp. 794--797, 1995.

\bibitem{geo}
R.~Parekh, J.~Yang, and V.~Honavar, ``Constructive neural-network learning
  algorithms for pattern classification,'' \emph{IEEE Transactions on neural
  networks}, vol.~11, no.~2, pp. 436--451, 2000.

\bibitem{yang1999distal}
J.~Yang, R.~Parekh, and V.~Honavar, ``Distal: An inter-pattern distance-based
  constructive learning algorithm,'' \emph{Intelligent Data Analysis}, vol.~3,
  no.~1, pp. 55--73, 1999.

\bibitem{marchand1989learning}
M.~Marchand, ``Learning by minimizing resources in neural networks,''
  \emph{Complex Systems}, vol.~3, pp. 229--241, 1989.

\bibitem{kuo2016understanding}
C.-C.~J. Kuo, ``Understanding convolutional neural networks with a mathematical
  model,'' \emph{Journal of Visual Communication and Image Representation},
  vol.~41, pp. 406--413, 2016.

\bibitem{huang2006extreme}
G.-B. Huang, Q.-Y. Zhu, and C.-K. Siew, ``Extreme learning machine: theory and
  applications,'' \emph{Neurocomputing}, vol.~70, no. 1-3, pp. 489--501, 2006.

\bibitem{huang2006universal}
G.-B. Huang, L.~Chen, C.~K. Siew \emph{et~al.}, ``Universal approximation using
  incremental constructive feedforward networks with random hidden nodes,''
  \emph{IEEE Trans. Neural Networks}, vol.~17, no.~4, pp. 879--892, 2006.

\bibitem{huang2007convex}
G.-B. Huang and L.~Chen, ``Convex incremental extreme learning machine,''
  \emph{Neurocomputing}, vol.~70, no. 16-18, pp. 3056--3062, 2007.

\bibitem{huang2008enhanced}
------, ``Enhanced random search based incremental extreme learning machine,''
  \emph{Neurocomputing}, vol.~71, no. 16-18, pp. 3460--3468, 2008.

\bibitem{chen2018saak}
Y.~Chen, Z.~Xu, S.~Cai, Y.~Lang, and C.-C.~J. Kuo, ``A saak transform approach
  to efficient, scalable and robust handwritten digits recognition,'' in
  \emph{2018 Picture Coding Symposium (PCS)}.\hskip 1em plus 0.5em minus
  0.4em\relax IEEE, 2018, pp. 174--178.

\bibitem{chen2020pixelhop}
Y.~Chen and C.-C.~J. Kuo, ``Pixelhop: A successive subspace learning (ssl)
  method for object recognition,'' \emph{Journal of Visual Communication and
  Image Representation}, vol.~70, p. 102749, 2020.

\bibitem{chen2020pixelhop++}
Y.~Chen, M.~Rouhsedaghat, S.~You, R.~Rao, and C.-C.~J. Kuo, ``Pixelhop++: A
  small successive-subspace-learning-based (ssl-based) model for image
  classification,'' in \emph{2020 IEEE International Conference on Image
  Processing (ICIP)}.\hskip 1em plus 0.5em minus 0.4em\relax IEEE, 2020, pp.
  3294--3298.

\bibitem{kuo2018data}
C.-C.~J. Kuo and Y.~Chen, ``On data-driven saak transform,'' \emph{Journal of
  Visual Communication and Image Representation}, vol.~50, pp. 237--246, 2018.

\bibitem{kuo2019interpretable}
C.-C.~J. Kuo, M.~Zhang, S.~Li, J.~Duan, and Y.~Chen, ``Interpretable
  convolutional neural networks via feedforward design,'' \emph{Journal of
  Visual Communication and Image Representation}, 2019.

\end{thebibliography}

\end{document}